\documentclass[letterpaper, 10 pt, conference]{ieeeconf}  

\IEEEoverridecommandlockouts                              

\usepackage[loading]{tracefnt}
\usepackage{graphics} 
\usepackage{fancyhdr,amsmath,amssymb,amsfonts,amsxtra,bm, hyperref}
\usepackage{dsfont}
\usepackage{algorithm,algorithmic}
\usepackage{graphicx}
\usepackage{comment,threeparttable}
\usepackage[table,xcdraw]{xcolor}
\DeclareMathOperator*{\argmin}{arg\,min}
\DeclareSymbolFont{matha}{U}{matha}{m}{n}
\DeclareMathSymbol{\Lt}{3}{matha}{"CE}
\DeclareMathSymbol{\Gt}{3}{matha}{"CF}
\usepackage{mathtools}
\definecolor{blue}{RGB}{0,0,0}

\newlength\myindent
\newcommand\bindent{%
  \begingroup
  \setlength{\itemindent}{\myindent}
  \addtolength{\algorithmicindent}{\myindent}
}
\newcommand\eindent{\endgroup}

\title{\LARGE \bf
Auditory cueing strategy for stride length and cadence modification: a feasibility study with healthy adults
}

\author{Tina LY Wu\,$^{1}$, Anna Murphy\,$^{2}$, Chao Chen\,$^{1}$, and Dana Kuli{\'c}\,$^{1*}$
\thanks{*This work was supported by the Natural Sciences and Engineering Research Council of Canada (NSERC), Monash Institute of Medical Engineering (MIME), The Commonwealth Scientific and Industrial Research Organisation (CSIRO), and Monash Partners. D. Kuli\'c was supported by an Australian Research Council Future Fellowship (FT200100761).}
\thanks{$^{1}$Tina, Chao \& Dana are with the Faculty of Engineering, Monash University, Victoria, Australia
        {\tt\small lee.wu@monash.edu, chao.chen@monash.edu, dana.kulic@monash.edu}}%
\thanks{$^{2}$Anna is with the Clinical Research Centre for Movement Disorders and Gait, Monash Health, Victoria, Australia
        {\tt\small anna.murphy2@monashhealth.org}}%
}

\begin{document}

\maketitle
\thispagestyle{empty}
\pagestyle{empty}

\begin{abstract}
People with Parkinson's Disease experience gait impairments that significantly impact their quality of life. Visual, auditory, and tactile cues can alleviate gait impairments, but they can become less effective due to the progressive nature of the disease and changes in people's motor capability. In this study, we develop a human-in-the-loop (HIL) framework that monitors two key gait parameters, stride length and cadence, and continuously learns a person-specific model of how the parameters change in response to the feedback. The model is then used in an optimization algorithm to improve the gait parameters. This feasibility study examines whether auditory cues can be used to influence stride length in people without gait impairments. The results demonstrate the benefits of the HIL framework in maintaining people's stride length in the presence of a secondary task. 
\newline

\indent \textit{Clinical relevance}— This paper proposes a gait rehabilitation framework that provides a personalized cueing strategy based on the person's real-time response to cues.  The proposed approach has potential application to people with Parkinson's Disease.
\end{abstract}

\section{INTRODUCTION}

Parkinson’s Disease (PD) is a progressive neurological disorder that affects movement. In advanced stages of the disease, a common lower body symptom is Freezing of Gait (FoG), where a breakdown of the person's existing stride-length-cadence relationship (SLCrel) occurs. In unimpaired gait, an increase in cadence (i.e. steps per minute) is typically accompanied by an increase in stride length (i.e. the distance travelled by the same foot) \cite{egerton2011central}. SLCrel can exhibit a positive linear, negative linear, or negative quadratic relationship  (ibid.). When participants walk at different self-selected speeds, 90-100\% of the participants across different age groups exhibit a SLCrel (ibid.). A freezing episode manifests as an abnormal increase in cadence with a noticeable decrease in step length, i.e. the SLCrel breaks down \cite{sweeney2019technological}. 

The use of wearable devices to monitor the person's gait, combined with feedback mechanisms in the form of visual, auditory, or tactile cues, has been shown to help people alleviate FoG \cite{sweeney2019technological}. Existing research on cueing mechanisms focuses on providing on-demand cues, where visual cues at a fixed distance or auditory/tactile cues at a fixed pace are provided at the onset of freezing \cite{sweeney2019technological, bachlin2010wearable, mikos2019wearable, schaefer2014auditory}. However, few studies have focused on adapting the provision of cues. Currently, adjustments to cues are performed by therapists during clinic visits. The lack of cue adaptation can decrease the cue effectiveness, given the day-to-day symptom variability and longitudinal disease progression. For instance, people can respond to cues differently due to changes in motor capability as part of the medication cycle \cite{sweeney2019technological} or may experience a decrease in responsiveness due to habituation \cite{ginis2018cueing}. 

In this study, we extend our previous work in \cite{me, me2} that proposes a cue-adaptation method based on the individual's real-time response to cues to address symptom variability and habituation. Previous cue-adaptation methods (e.g. \cite{zhang2020wearable, zhang2022reinforcement}) focus on increasing gait speed (m/s), which is influenced by both stride length and cadence. However, the studies assume the change in gait speed will positively influence step length and cadence. The assumption can be detrimental in Parkinson's as a faster gait speed can also be achieved by increasing cadence, which could trigger FoG \cite{spaulding2013cueing}. Our current work explicitly models the SLCrel and provides cues that account for both gait parameters. In our previous work, we demonstrated that auditory cues can influence cadence in healthy adults \cite{me} and people with PD \cite{me2}. This feasibility study focuses on the stride length aspect of the SLCrel. The evaluation is important as previous studies have suggested that auditory cues are not always effective in changing stride length \cite{suteerawattananon2004effects, spaulding2013cueing}. Results from this study will inform whether a change in cueing modality is needed and provide an initial evaluation of the proposed HIL framework.

\section{METHODS}
\subsection{Proposed Framework}
\begin{figure}[!t]
\centering
\includegraphics[width=\columnwidth]{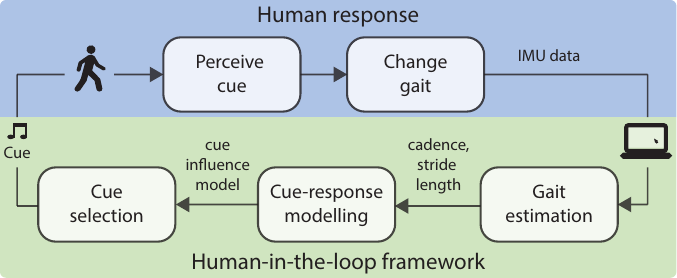}
\caption{The block diagram of the HIL framework with its user.}
\label{fig:block-diagram}
\end{figure}

We extend our previous work in \cite{me} to incorporate monitoring of the SLCrel. The HIL framework consists of three sub-components: an online gait parameter estimation algorithm, a model of cue influence on gait, and an optimization function for cue provision. The framework is illustrated in Figure \ref{fig:block-diagram}.

\subsubsection{Estimate gait parameters online}\label{sec:gait_estimation}
A key requirement of the HIL framework is to measure stride length in real-time. The system needs to have fast computation and be low-cost, portable, and easy to set up. Based on these requirements, a solution using two IMU sensors secured onto each foot is implemented. Previously, the zero-velocity update (ZVU) algorithm \cite{skog2010zero} had been developed to correct for sensor drift during the stance phase of the gait cycle. To apply ZVU, we extend the method from \cite{van2016real}, where a dynamic threshold is computed from the accelerometer and gyroscope signals to distinguish between the stance and swing phases. The method is person-invariant and speed-invariant, and reduces experiment complexity (i.e. no need to tune for each participant) and computation requirements (i.e. no need to train on a large amount of data or need expensive hardware to run). One benefit of real-time stance/swing detection is being able to synchronize cues to a gait phase, similar to others such as \cite{mancini2018assessment}. In addition, the detection can be used to estimate cadence (the time of the step from the start of one swing phase to the next). However, \cite{van2016real} did not perform sufficient validation with participants. We found the algorithm was not robust against interpersonal variations and would often lead to false positives. Therefore, we augmented the method and summarized it in Algorithm \ref{alg:detect_swing}. The algorithm takes each IMU sample ($\textbf{a}^{t+1}, \bm{\omega}^{t+1}, \textbf{q}^{t+1}$ for accelerometer, gyroscope, and quaternion) at the current time step, $t+1$, and \textcolor{blue}{returns true if it is greater than the dynamic threshold ($th_{dynamic}$). $th_{dynamic}$ is a function of the aggregated features computed from the IMU samples and it is the key to the person and speed-invariant algorithm.} The swing classification result is stored in the array called $is\_swing$. Two additional lists, called $rising\_edge\_list$ and $falling\_edge\_list$ are added as secondary filters to reject false positives.

\begin{algorithm}
\caption{Swing detection algorithm}
\label{alg:detect_swing}
\begin{algorithmic}

    \STATE Function \textbf{DetectSwing}($\textbf{a}^{t+1}, \bm{\omega}^{t+1}, \textbf{q}^{t+1}$)
        \bindent
            \STATE Update rolling window 
            \STATE Compute features from the window  
            \STATE Aggregate features from above

            \IF{Aggregated feature $>$ \textcolor{blue}{$th_{dynamic}$}}
                \STATE $is\_swing^{t+1} \gets True $
                \STATE Initialize debounce counter
            \ELSE{
                \STATE $is\_swing^{t+1} \gets False $
            }
            \ENDIF

            \IF{Aggregated feature $<$ \textcolor{blue}{$th_{static}$}}
                \STATE $is\_swing^{t+1} \gets False $
                \STATE Increment debounce counter
            \ENDIF

            \IF{$is\_swing^{t+1}$ and waiting for a rising edge}
                \STATE Append $t+1$ to $rising\_edge\_list$ 
            \ENDIF

            \IF{$not\text{ } is\_swing^{t+1}$ and waiting for a falling edge and is debounced}
                \STATE Update step-specific features
                \STATE \textcolor{blue}{Update $th_{dynamic}$ \& $th_{static}$}
                \STATE Append $t+1$ to $falling\_edge\_list$ 
            \ENDIF

        \eindent
\end{algorithmic}
\end{algorithm}

\begin{figure}[!t]
\centering
\includegraphics[width=\columnwidth]{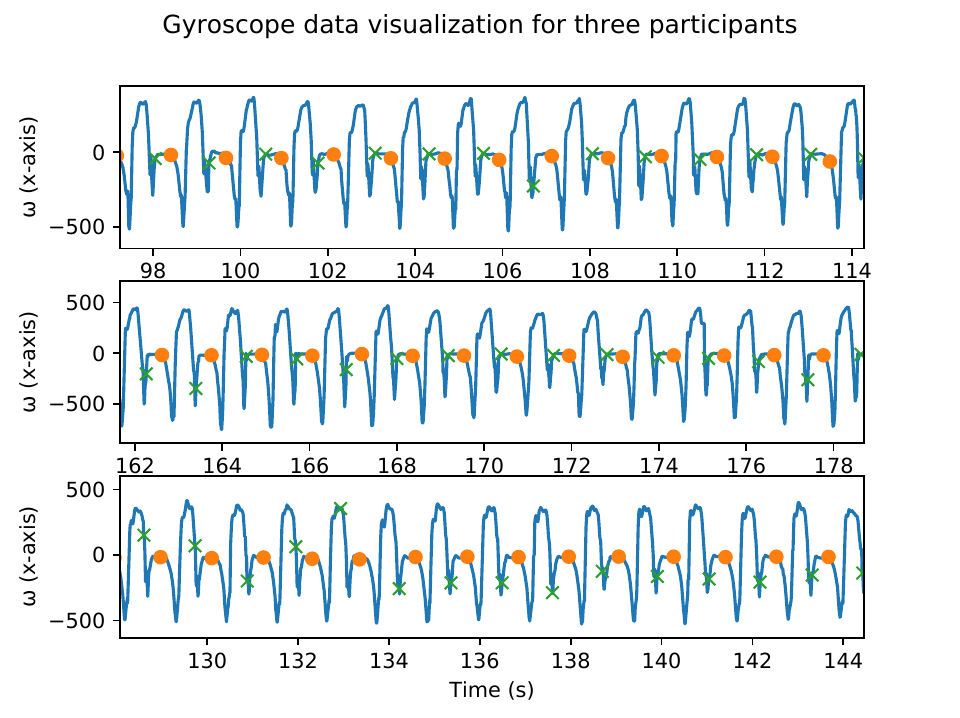}
\caption{The gyroscope data in the x-axis from the sensor located on the dominant leg of 3 participants during the experiment. The walking path contains both straight lines and turns. The orange dot indicates the start of the swing phase, and the green x indicates the end. The start detection is consistent, but the end-of-swing can sometimes be premature (as shown in the third sub-figure, where x is labelled when the foot is yet to become stationary). This is consistent with our Vicon validation described in Section \ref{sec:gait_estimation}, where the STD is larger during turns compared to straight-line walking.}
\label{fig:gyro-data}
\end{figure}

Examples of the algorithm performance are shown in Figure \ref{fig:gyro-data}. Once the stance/swing phases are detected, we applied the algorithm described in \cite{madgwick2010efficient} by adapting the code from \href{https://github.com/xioTechnologies/Gait-Tracking-With-x-IMU}{xioTechnologies/Gait-Tracking-With-x-IMU}. The full implementation is provided in the Appendix. We conducted a validation study for the stride length estimation algorithm with two healthy adult participants using Cometa System IMUs and the Vicon motion capture system. The validation study is divided into both straight-line walking and circle walking due to the limited size of the Vicon rig. Both walking patterns are important as our planned path for the experiment requires both motions, with the circle walking representing turns around the corners of the room. The percent error and standard deviation (STD) for straight-line walking is -0.09$\pm$0.03 and the error for circle-walking is -0.024$\pm$0.19 meters. 

\subsubsection{Model cue influence on gait}
A sparse multi-output Gaussian Process (MOGP) is used to model the change in gait parameters (i.e. cadence and stride length) as a result of a given auditory cue. The model takes the form of $f(x): \mathds{R}^D \rightarrow \mathds{R}^P$, where D is the dimension of the input and P is the dimension of the output. 

\begin{equation}
\textbf{Y} = f(\textbf{x}) = \mathrm{W}\textbf{g}(\textbf{x}), \label{eq:gp}
\end{equation}

\noindent $\textbf{g}(\textbf{x})$ is a collection of Q independent GPs, where Q is the number of latent GPs:

\begin{equation}
    \textbf{g}(\textbf{x}) = \{ g_q(\textbf{x}) \}_{q=1}^Q, g_q(\cdot) \sim \mathcal{GP}(0, k_q(\cdot,\cdot'))
\end{equation}

\noindent The outputs are assumed to be linearly correlated through $\mathrm{W}$, known as the Linear Model of Coregionalization (LMC) following the implementation specified in \cite{van2020framework}. The outputs of the MOGP are the estimated cadences and stride lengths, while the inputs are the specified cues at the preceding time step:

\begin{align}
&\textbf{Y} = \begin{bmatrix}
\hat{f}_1, \hat{\ell}_{1} \\
\hat{f}_2, \hat{\ell}_{2}  \\
\vdots \\
\hat{f}_N, \hat{\ell}_{N}
\end{bmatrix} \notag
&\textbf{x} = \begin{bmatrix}
0 \\
c_1 \\
\vdots \\
c_{N-1}
\end{bmatrix} \notag
\end{align}

\noindent $\hat{f}_n$ is the estimated cadence and $\hat{\ell}_{n}$ is the estimated stride length at the $n^{th}$ step from the gait measurement sub-system. $c_{n-1}$ is the cue given at the previous step that results in the $n^{th}$ cadence/stride length. \textit{n} is incremented at every footstep and $n = [1, 2, \hdots, N]$. $n=1$ represents the baseline cadence and stride length when no cue is given.

A challenge associated with MOGP is the computation complexity associated with the covariance matrix manipulation, which grows cubically with respect to the amount of data. \cite{alvarez2010efficient} has proposed using variational free energy approximation combined with inducing points to construct a sparse approximation that reduces the computation cost. The result is implemented in a python library (GPFlow) \cite{van2020framework}, which is utilized in this study. Specifically, sparsity is introduced to the MOGP through inducing points, $\pmb{\mathrm{Z}} = [z_1, z_2, ... z_M]$. Then, the MOGP prior, $p_0$, can be written in terms of $\pmb{\mathrm{Z}}$, where

\begin{equation}
    p_0({\textbf{g}_q}) = \mathcal{N}(m_q(\textbf{Z}), k_q(\textbf{Z},\textbf{Z}'))
\end{equation}

The key model parameters are chosen as the following: $P=Q=2, D=1, M=20$. The covariance, $k_q(\cdot,\cdot')$, is chosen to be the sum of a squared exponential kernel and a constant kernel, which allows possible higher order SLCrel to be captured. To use the model to predict change in stride length and cadence as a result of the input cues, the model is evaluated at the new input location following Eq~\ref{eq:gp_pred}.

\begin{equation}\label{eq:gp_pred}
    Y^{\star} = Wg(x^\star)\text{,}\quad x^\star = [c^{\star}_n]\text{,}\quad Y^\star = [\hat{f}^{\star}_{n+1}, \hat{\ell}^{\star}_{n+1}] 
\end{equation}

\subsubsection{Cue Optimization}
The cue-optimizing sub-system aims to provide cues that minimize the difference between the predicted and desired gait states. The cost function penalizes the squared difference between target cadence/stride length and predicted cadence/stride length, as well as rapid cue changes.

\begin{gather}
c_{opt} = \argmin_{c^\star_{n}} J \mbox{, subjected to } c_{min} \leq c^\star_{n}\leq c_{max} \notag \\
J(c^\star_{n}) = \alpha_{f} ( f_{target} - \hat{f}^{\star}_{n+1})^{2} + \alpha_{l}( \ell_{target} - \hat{\ell}^{\star}_{n+1}) ^2 \notag \\
+ \alpha_{e}(c^\star_{n} - c_{n-1})^2 \label{eq:cost}
\end{gather}

\noindent In Eq~\ref{eq:cost}, $c_{opt}$ is the cue to provide at the $n+1^{th}$ step constrained between $c_{min}=0.65f_{baseline}$ and $c_{max}=1.35f_{baseline}$. The range is determined empirically based on a previous study \cite{me}. $\alpha_{f}, \alpha_{l}, \text{and } \alpha_{e}$ are three scaling factors that weigh the relative importance of each cost term, which are initialized to 1.5, 10, and 0.05 respectively. $\hat{f}^{\star}_{n+1}$ is the predicted (i.e. $n+1$ step) cadence and $\hat{\ell}^{\star}_{n+1}$ is the predicted stride length estimated from the MOGP given the cue at the current step, $c_n^\star$, using Eq~\ref{eq:gp_pred}. The cost function is solved using the Nelder-Mead method in SciPy \cite{2020SciPy-NMeth}.

\subsection{Target Selection} \label{sec:target_selection}
\begin{figure}[!t]
\centering
\includegraphics[width=\columnwidth]{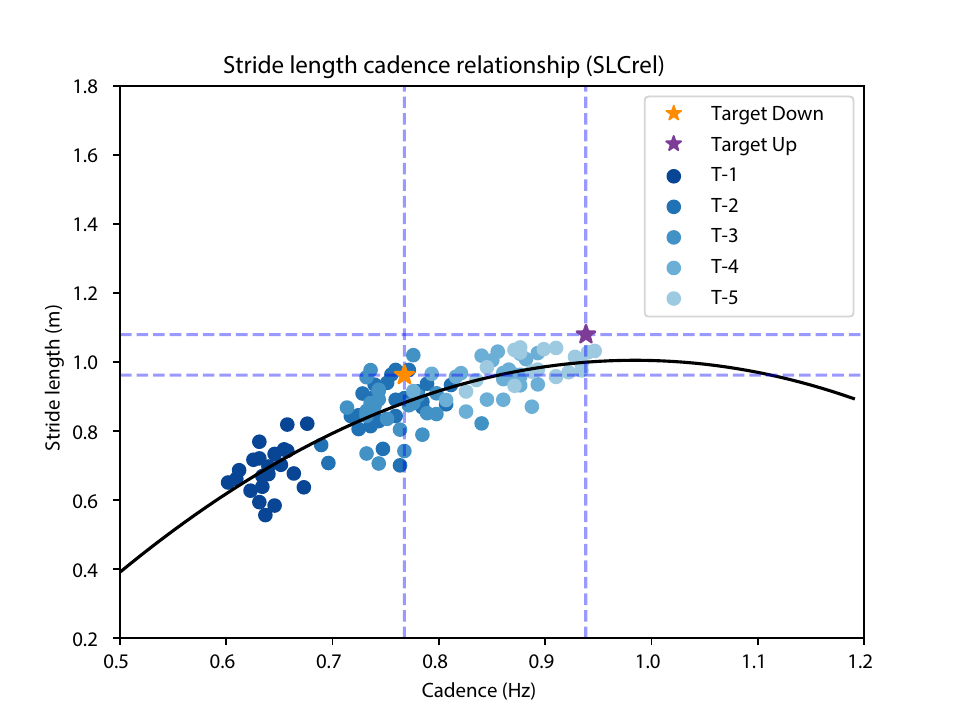}
\caption{Example of the initial SLCrel. The dots represent the stride length/cadence data collected during the training beats (labelled from T-1 to T-5 from slow to fast). The participant exhibits a negative parabolic SLCrel and $f_{target} \& \ell_{target}$ are chosen to be the location of the lighter orange star labelled ``Target Up''.}
\label{fig:slcrel}
\end{figure}

$f_{target}$ and $\ell_{target}$ in Eq \ref{eq:cost} are selected based on the participant's baseline cadence ($f_{baseline}$) and initial SLCrel measured at the start of the experiment. $f_{baseline}$ is measured in a 6-Minute Walk Test (6MWT) and the SLCrel is measured by providing the participants with 5 training beats at 1.16, 1.41, 1.58, 1.75, 1.91 Hz in a random order (values adopted from \cite{egerton2011central}). 50 beats are provided for each frequency. A quadratic and a linear polynomial is fitted to the training data using NumPy \cite{schwartze2011impact} and the polynomial with a lower residual becomes the SLCrel. An example of SLCrel is shown in Figure \ref{fig:slcrel}. $\ell_{target}$ is selected to be a 0.1 m increase from the SLCrel, which is based on the error of the gait estimation algorithm. Two candidate targets are evaluated by computing the y-values at $\pm10\%f_{baseline}$ on the SLCrel and adding a 0.1 offset. The exact target is chosen by looking at the two candidates $\ell_{target}$ and the higher value is chosen. $f_{target}$ is then selected (i.e. the x value of $\ell_{target}$ on the SLCrel). 

\subsection{Experimental Conditions}\label{sec:exp_cond}
We compare the performance of two cueing strategies in two scenarios, leading to 4 experimental conditions. Each condition lasts for 4 minutes. Since healthy participants do not experience gait impairments, we evaluate the performance of the framework in terms of its ability to change people's stride length. The two cueing strategies are the fixed and the adaptive strategy. The fixed strategy delivers cues directly at $f_{target}$ and the adaptive strategy provides cues using the HIL framework. We initialize the MOGP using the data from the initial SLCrel. The two scenarios are either with or without a secondary task. For the secondary task, participants perform a word reciting task (i.e. reciting as many words beginning with a given letter). A set of letters (S, P, C, and A), are randomly selected before the experiment and randomly assigned. Conditions with secondary task are designed to emulate natural daily living where people could be preoccupied with other tasks while walking. 

\subsection{Participants}
We recruited 6 healthy adults (5M/1F, Age $27.5\pm3.78$ years; Height $172.17\pm6.4$ (cm); Weight $71.67\pm8.11$ (kg); mean$\pm$standard deviation). We started the experiment with three participants (Group 1) who only received 1 training session (described below) and the cue-triggering conditions are described in Eq \ref{eq:sl_trigger} \& \ref{eq:cad_trigger} over a 5-step window ($n_{window}$): 

\begin{equation}\label{eq:sl_trigger}
\dfrac{\sum^{N}_{n=N-4} \hat{\ell}_{n}}{ n_{window}} < \ell_{target} 
\end{equation}
\begin{equation}\label{eq:cad_trigger}
    \text{ or } \dfrac{ \dfrac{\sum^{N}_{n=N-4} \hat{f}_{n}}{ n_{window}} - f_{target}} {f_{target}} \leq 5\%
\end{equation}

We relaxed the cue-triggering condition for the next three participants (Group 2) to use only Eq \ref{eq:sl_trigger} (see Section \ref{sec:discussion} for relevant discussion).

\subsection{Protocol}\label{subsec:protocol}
Participants watched an introduction video and signed the consent form at the start of the experiment. After putting on the IMUs, participants were given the first training session where a metronome beat is randomly selected by the experimenter and participants practice syncing their walking to the metronome beat. \textcolor{blue}{The training takes less than one minute.} Participants walked around a room, with the longest straight edge being 15 meters. The walking path contained both straight-line walking and 4 corner turns. Participants were free to choose the direction (clockwise or counter-clockwise) of their walk and kept to the same direction throughout. After the training session, participants were told to walk at their natural pace for the 6MWT. After the 6MWT, participants filled out a demographic survey. The initial SLCrel is then constructed (see Section \ref{sec:target_selection}). Participants were told to sync their walking to the metronome beat for each condition. 

For the first three participants (Group 1), the next part of the experiment involved the 4 experimental conditions described in Section \ref{sec:exp_cond}, which were selected randomly and blinded from the participants. Participants filled out a survey plus the NASA Task Load Index (TLX) after each condition.

For the next three participants (Group 2), a second training session, \textcolor{blue}{which takes approximately one minute}, was provided before the experimental conditions. In the training, the experimenter first played another randomly selected metronome beat and walked with the participant. After a few steps of syncing to the beat, the experimenter asked the participant to ``take bigger steps'' while keeping to the same beat. The experimenter then asked the participants to ``take smaller steps'' while keeping to the same beat. The participants were told that the training aims to demonstrate how various step lengths can be associated with a beat. Participants were then asked to try and figure out the intention of the framework during the experiment in terms of how fast and how far the metronome wanted them to step. They would know they have it correct when the beats turn off, and the goal was to keep the beats off. 

Participants in both groups were told they will be evaluated based on their gait performance and the number of words they can recite if the secondary task is present. All participants concluded the experiment after an interview during which the participant reviewed their own data and the experimenter informed the participants of the cueing strategies. The study (ID 34903) was approved by the Monash University Human Research Ethics Committee.

\subsection{Materials}
The IMU sensors from the WaveTrack Inertial System  are sampled at 142 Hz (Cometa Systems, Milan, IT) and streamed wirelessly into a custom Python program. The program runs on a laptop (Windows 10, i7 core with no GPU), which plays auditory cues from a speaker (Phillips BT50A). Two sensors are used; one on each foot. The sensor is fixed to the flat part of the foot, which is identified by asking participants to lift their heels; the sensor is then placed on top of the folding crease and secured using duct tape. The sensor is oriented such that the x-y plane of the sensor is parallel to the transverse plane of the body. The sensor's y-axis points in the sagittal plane facing forward and z-axis points towards the head.

\subsection{Analysis}
No statistical analysis was conducted due to the small sample size. Here we focus on the main gait metric (stride length) and participants' subjective ratings.

\section{RESULTS} \label{sec:results}
\begin{figure*}[htb]
\begin{center}
\includegraphics[width=\textwidth]{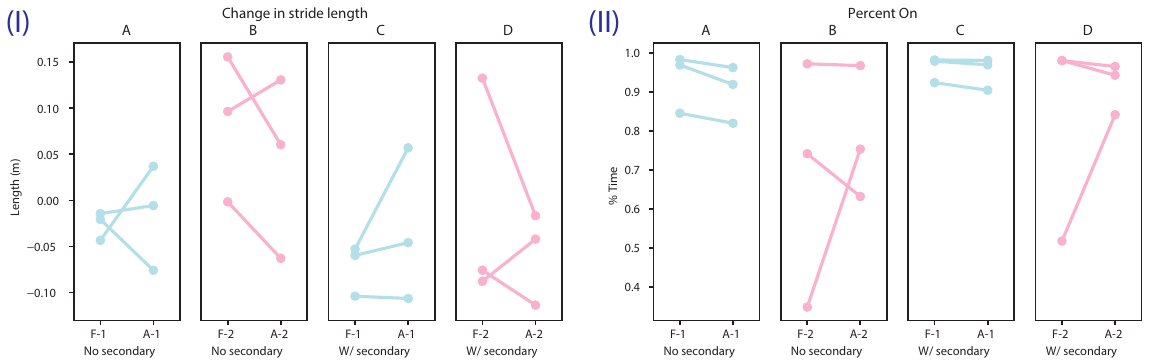}
\caption{The figure displays two outcome metrics of the study. (I) shows the change in stride length compared to baseline stride length. We want the metric to be as high as possible (i.e. longer steps). (II) shows the percentage of time the cue is on during the 4-minute experiment. The goal is for the percent on time to be low. The connected lines between two columns show how the individual's data changed between the fixed and adaptive approaches. The two left panels (A) \& (B) are the conditions without secondary task and (C) \& (D) are with the task. F refers to the fixed cueing strategy and A refers to the adaptive strategy. The 1 in the x-axis denotes the groups. Group 1 received no additional training and is subjected to both $f_{target}\&\ell_{target}$. Group 2 received 2 training sessions and only had the $\ell_{target}$ active for cue-triggering.}
\label{fig:agg-data}
\end{center}
\end{figure*}

This feasibility study aims to determine whether participants are able to increase their stride length using auditory cues. We calculate the delta stride length, which is the mean difference between the participant's stride length during a condition compared to the mean of their baseline stride length during the 6MWT. The result is shown in Figure \ref{fig:agg-data}. Data in Figure \ref{fig:agg-data}(I)-A indicate that participants from Group 1 are not changing their strides even during the condition without the secondary task. Without sufficient information, the cue is delivered inefficiently as it is played almost 100\% of the time without being able to influence the participant's gait, as shown in Figure \ref{fig:agg-data}(II)-A\&C. Despite the lack of instructions, the adaptive condition (A-1) naturally encourages participants to explore a variety of step lengths through the change in cues (as evidenced by the larger variance in the data) as seen in Figure \ref{fig:agg-data}(I)-A. With more exploration, participants had a higher chance of meeting the cue-triggering conditions described in Eq \ref{eq:cad_trigger} \& Eq \ref{eq:sl_trigger}, thereby turning the metronome off (i.e. Figure \ref{fig:agg-data}(II)-A, \textcolor{blue}{where the lines between F1 and A1 trends downwards, favouring the adaptive approach}). When the secondary task is added without sufficient instruction, further reduction in step length is observed (Figure \ref{fig:agg-data}(I)-C$<$A). However, the adaptive condition still encouraged more variations in step length, leading to a closer-to-baseline median (seen in Figure \ref{fig:agg-data}(I)-C, \textcolor{blue}{where two participants trends upwards between F-1 and A-1, favouring the adaptive approach. One participant has a slight reduction}). 

Once the instructions are modified for Group 2, participants started changing their stride length, as evident by the positive increase in step length in Figure \ref{fig:agg-data}(I)-B. A larger change in step length in the fixed approach than in the adaptive approach is observed in the conditions without secondary task (\textcolor{blue}{lines trend downward in Figure \ref{fig:agg-data}(I)-B between F-2 and A-2, favouring the fixed condition}). \textcolor{blue}{This could be due to participants varying their step lengths when beats at a new pace are provided, which lowers the overall change in step length}. \textcolor{blue}{Overall, the stride length decreases in the presence of the secondary task (Figure \ref{fig:agg-data}(I)-B$>$D). The median percent on time is lower without secondary task (Figure \ref{fig:agg-data}(II)-B$<$D). When the secondary task is added, two participants experienced a decrease in percent on time compared to the fixed approach and one increased drastically (Figure \ref{fig:agg-data}(II)-D)}.


\textcolor{blue}{The initial results first suggest that without explicit association between stride length and cues (Group 1), the adaptive approach is more effective in changing people's stride length (as most lines trend upwards between fixed and adaptive in Figure \ref{fig:agg-data}(I)-A\&C) reducing the percent on time (as most lines trend downwards between fixed and adaptive in Figure \ref{fig:agg-data}(II)-A\&C). When the second training session is added (Group 2), the results suggest there could be two ways of responding to the adaptive approach. For some participants, the adaptive approach continues to improve stride length and reduces percent on time, while others are confused by the non-static nature of the cue. This is supported by the post-condition questionnaires}. When asked if and how the participants felt their gait changing, Group 2 answered that their gait is ``random'' more often than Group 1. This suggests that participants in Group 2 are consciously exploring the variations in stride length, but may still be difficult for them to ``figure out'' the intention of the framework as per instruction. This is also supported by the Task Load Index (TLX) score questions, where the largest and second largest change is seen in the participant's stress/irritation level (i.e. frustrated about not being able to turn off the metronome) followed by the physical demand of the task (i.e. exploring more step lengths). Overall, TLX increased between Group 1 and Group 2 (16.58$\pm$7.12 for Group 1 and 18.18$\pm$6.22 for Group 2; mean$\pm$standard deviation).

\section{DISCUSSION}\label{sec:discussion}
In this study, we demonstrate that it is possible to change people's stride length using auditory cues given sufficient instructions. This aligns with how cues are used during rehabilitation by bringing attention to the walking task. For Parkinson's, the attentional mechanism helps bypass the defective automatic control due to the disease, thereby improving their walking \cite{sweeney2019technological}. In the modified instruction (i.e. with Group 2), we emphasize attention control, which resulted in a greater change in stride length and higher TLX scores. In addition to the instruction, the cue-triggering conditions described in Eq \ref{eq:sl_trigger} and \ref{eq:cad_trigger} could also contribute to participants' lack of stride changes. This is because the $\ell_{target}$ that is selected is not part of the natural SLCrel, and therefore satisfying both conditions may not be feasible. Finally, since $f_{target}$ can easily be satisfied (as a participant put it: ``[I] simply match the pace with the metronome''), a change in stride length in order to keep the metronome off was difficult to realize due to the natural association between walking frequency and metronome frequency. Therefore, the demonstration session was necessary to highlight the connection between metronome frequency and stride length. 

The initial results for Group 2 suggest that the fixed approach performs better without secondary task, but the adaptive approach is better when the secondary task is added, given the reduction in percent on time. In addition, participants mentioned that the adaptive approach in general is more attention-demanding compared to the fixed approach. From these results, we plan to expand the study to include people with Parkinson's as well as a control group of older adults to evaluate the performance of the cueing strategies in relation to the SLCrel.


\section*{APPENDIX}
The full implementation of the gait detection algorithm and instructions on parameter tuning can be found here: \href{https://doi.org/10.26180/c.6619669.v3}{https://doi.org/10.26180/c.6619669.v3}.


\bibliographystyle{IEEEtran}
\bibliography{reference}

\end{document}